# Multi-centric AI Model for Unruptured Intracranial Aneurysm Detection and Volumetric Segmentation in 3D TOF-MRI

Ashraya K. Indrakanti MD[1,2], Jakob Wasserthal PhD[2], Martin Segeroth MD[2], Shan Yang PhD[2], Victor Schulze-Zachau MD[1,2], Joshy Cyriac MSc[2], Michael Bach PhD[2], Marios Psychogios MD[1,2]*, Matthias A. Mutke MD[1,2]*

*Both authors contributed equally and share last authorship.

[1] Department of Diagnostic and Interventional Neuroradiology, University Hospital Basel, Basel, Switzerland
[2] Clinic of Radiology and Nuclear Medicine, University Hospital Basel, Basel, Switzerland

Corresponding Author: matthias.mutke@usb.ch

## Abstract

**Purpose:**
To develop an open-source nnU-Net-based AI model for combined detection and segmentation of unruptured intracranial aneurysms (UICA) in 3D TOF-MRI, and compare models trained on datasets with aneurysm-like differential diagnoses.

**Methods:**
This retrospective study (2020-2023) included 385 anonymized 3D TOF-MRI images from 364 patients (mean age 59 years, 60% female) at multiple centers plus 113 subjects from the ADAM challenge. Images featured untreated or possible UICAs and differential diagnoses. Four distinct training datasets were created, and the nnU-Net framework was used for model development. Performance was assessed on a separate test set using sensitivity and False Positive (FP)/case rate for detection, and DICE score and NSD (Normalized Surface Distance) with a 0.5mm threshold for segmentation. Statistical analysis included chi-square, Mann-Whitney-U, and Kruskal-Wallis tests, with significance set at $p < 0.05$.

**Results:**
Models achieved overall sensitivity between 82% and 85% and a FP/case rate of 0.20 to 0.31, with no significant differences ($p = 0.90$ and $p = 0.16$). The primary model showed 85% sensitivity and 0.23 FP/case rate, outperforming the ADAM-challenge winner (61%) and a nnU-Net trained on ADAM data (51%) in sensitivity ($p < 0.05$). It achieved a mean DICE score of 0.73 and an NSD of 0.84 for correctly detected UICA.

**Conclusions:**
Our open-source, nnU-Net-based AI model (available at 10.5281/zenodo.13386859) demonstrates high sensitivity, low false positive rates, and consistent segmentation accuracy for UICA detection and segmentation in 3D TOF-MRI, suggesting its potential to improve clinical diagnosis and for monitoring of UICA.

# 1. Introduction

Unruptured Intracranial Aneurysms (UICA) affect approximately 3% of the population and pose a significant risk for subarachnoid hemorrhage upon rupture, associated with high morbidity and mortality rates (Vlak et al., 2011; Lantigua et al., 2015). Early detection and precise measurement of UICA size are critical for effective monitoring and potential treatment (Keedy et al., 2006) to prevent such debilitating outcomes.

For UICA assessment, 3D TOF-MRI is most used, primarily due to its noninvasiveness, absence of radiation, and lack of contrast agents (Keedy et al., 2006). Many UICAs are detected incidentally during routine clinical imaging for unrelated pathologies. Additionally, follow-up imaging to monitor UICA for prognostically relevant changes in size and shape is also usually conducted with 3D TOF-MRI.

Detecting and segmenting small UICAs in these images is challenging, with radiologists' sensitivity estimated between 60% and 85% (Lehnen et al., 2024; Sohn et al., 2021). However, comprehensive studies evaluating human error rates in routine imaging are lacking.

AI systems, particularly the nnU-Net, have shown significant potential in detection and segmentation tasks due to their self-configuring capabilities (Isensee et al., 2021). While multiple studies have explored various algorithms for UICA detection (Chen et al., 2023; Terasaki et al., 2021; Ueda et al., 2019; Lehnen et al., 2022), few have focused on a single model capable of detection and segmentation (Claux et al., 2023; Sichtermann et al., 2019). Moreover, we incorporate challenging and potentially confounding differential diagnoses, such as infundibula and vascular loops, in the training process. These are crucial for clinical accuracy, and their omission from training datasets might result in reduced sensitivity in clinical scenarios.

This study aims to develop an AI model for combined detection and 3D segmentation of UICAs in 3D TOF-MRI brain scans and to compare the performance of models trained on additional datasets containing potential confounding diagnoses. The primary objective was to assess the performance of the models in accurately detecting and segmenting confirmed UICA. We compared their performance to the performance of the ADAM-challenge winner (Timmins et al., 2021) and to an nnU-Net model trained solely on the ADAM-challenge dataset. The secondary objective was to compare the performances of these differently trained models and ascertain if significant variations existed among them. Such a model could enhance the accuracy of UICA detection and segmentation, ultimately reducing missed and false diagnoses, and aiding in the monitoring and therapy decisions for UICAs.

# 2. Materials and Methods

## 2.1. Data Acquisition

This retrospective study received an ethics waiver from the local institutional Review Board under project-ID Req-2024-00337.

We performed a retrospective, diagnostic imaging study using 385 randomly sampled, anonymized, non-contrast 3D TOF-MRI images acquired from 364 patients between 2020 and 2023 in a clinical imaging setting from our institute and its affiliates (Institutional Data). Additionally, we used the dataset from the ADAM challenge (Timmins et al., 2021). Further acquisition details are provided in the Supplement.

Inclusion criteria were studies containing one or more untreated UICA. Also, studies with UICA differential diagnoses (infundibula, vascular loops, fenestrations and focal ectasias) were included, i.e. the uncertainty of the radiologist prevented a definitive diagnosis between a true aneurysm or one of the differential diagnoses listed above. The data was then split into training (70%, n=242) and test data (30%, n=101). The only exclusion criterium for studies was the lack of sufficient imaging quality of the TOF-MRIs (see flowchart in Figure 1), resulting in the dataset containing diverse, non-aneurysmatic cerebral pathologies.

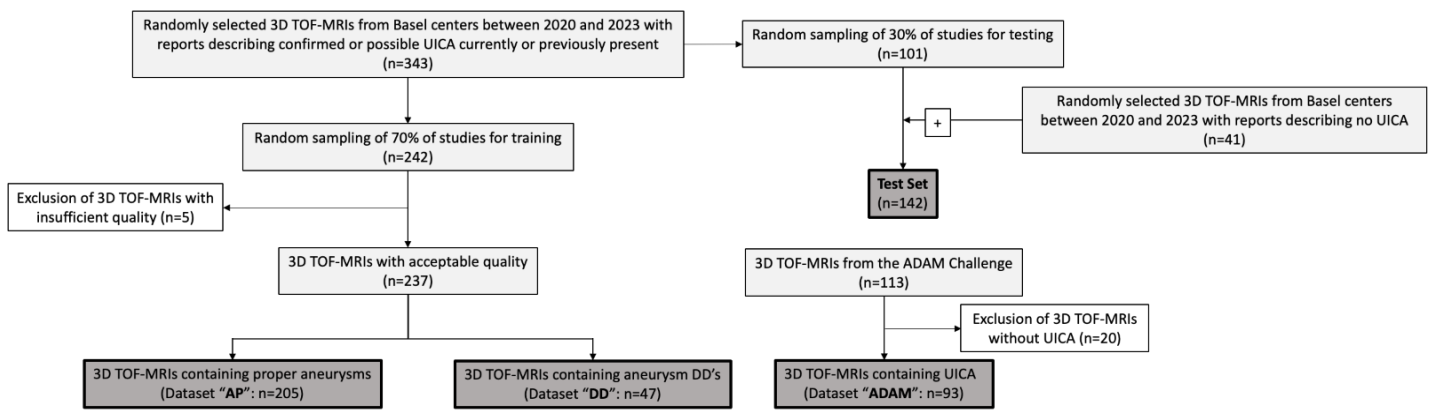

Figure 1. Study flowchart. After collection of 3D TOF-MRIs possibly containing UICA and excluding images not fulfilling inclusion criteria or with insufficient quality, the data was reorganized into four distinct datasets: AP, DD, ADAM and Test datasets. The first three datasets were used for model training in different combinations (AP, AP+DD, AP+ADAM, AP+DD+ADAM), while the Test dataset was used for model performance evaluation and model comparison.

The mean age of Institutional Data subjects was 59 years (range: 5-88 years), 60% female (female ages 9-87 years, mean 60 years; male ages: 5-88 years, mean 57 years). For the ADAM data, the median age was 55 years (range: 24 years to 75 years), 75% female (Timmins et al., 2021).

2.2) Data Categorization

We categorized the training data into three distinct training datasets for comprehensive analysis (Figure 1). The first dataset "Aneurysm Proper" (AP, n=205) included only studies where an UICA was diagnosed in the clinical reports. The second dataset "Aneurysm Differential Diagnoses" (DD, n=47) included studies where a UICA differential diagnosis was identified in the routine clinical reports. To introduce institutional variability, 93 out of 113 studies from the ADAM challenge (all positive studies containing at least one UICA) yielded the third "ADAM" dataset.

For the final test dataset, in addition to the 101 test studies containing only UICA and no UICA differential diagnoses, 41 randomly sampled 3D TOF-MRI images acquired between 2020 and 2023 with no reported UICA were added.

2.3) Data segmentation

The voxel-by-voxel segmentation of the Institutional Data studies was performed by a junior medical doctor (A.K.I.), guided by the corresponding radiology reports. This task was supervised by two board-certified neuroradiologists (M.M. with 10 years, and M.N.P. with 15 years of experience). In instances where the segmentation posed challenges, disagreements or deviated from initial radiology reports, a consensus was established among all parties involved. The ADAM-challenge dataset is pre-segmented.

In the AP dataset, all proper UICA were segmented. In the DD dataset, all UICA differential diagnoses were segmented. The manual segmentation and refinement of the preliminary segmentations was performed using the medical image editing software, NORA (Anastasopoulos et al., 2017).

2.4) Model Training and Evaluation

We utilized the nnU-Net framework (Isensee et al., 2021) to develop a model for detection and segmentation of UICA. We trained multiple nnU-Net models using combinations of the three datasets delineated in Figure 1 and described in Table 1, resulting in four distinct models (henceforth individually referred to as the AP model, AP+DD model, AP+ADAM model and AP+DD+ADAM model, and collectively referred to as IH models (In-House models)). The names of the models reflect the combinations of datasets the models were trained on.

Testing was performed on the separate test dataset described above. UICA were considered detected if there was any overlap (>0%) between the prediction and the ground truth lesion. Detection metrics included sensitivity and FP/case (the average number of false positive results per imaging study). The performance of the segmentations was evaluated using the lesion-wise DICE score and lesion-wise Normalized Surface Distance (NSD) with 0.5 mm threshold for all correctly detected UICA. This approach was used to simulate clinical practice, where a radiologist would only assess correctly detected UICAs for segmentation (analogous to Timmins et al., 2021). We extracted maximal diameter and volume of the UICA.

Using these metrics, the IH models were compared amongst each other, and compared to the open-source winning model of the ADAM challenge (JunMa11, 2020; henceforth referred to as the ADAM-winner model) and to a simple, standard nnU-Net model trained only on the ADAM dataset using default settings, henceforth called the ADAM-nnU-Net model.

2.5) Statistical Analysis

Detection was evaluated using the $\chi^2$-test. Mann-Whitney-U and Kruskal-Wallis tests were used to compare lesion sizes and segmentation accuracy among the different models. 95%-CI were calculated via nonparametric percentile Bootstrapping (N = 104). P-values < 0.05 were considered statistically significant. Hypothesis testing was two-tailed. Statistical Analysis was performed in Python (version 3.11) with SciPy (version 1.11.3) (Virtanen et al., 2020).

# 3. Results

3.1) Data

All UICA in the Institutional Data were saccular, with very few exceptions: Seven were fusiform and one was mycotic. Since the only exclusion criterium for studies was insufficient imaging quality, the dataset encapsulated a wide range of pathologies. Table 1 contains details regarding the number and sizes of UICA in the different datasets.

|  | AP | DD | ADAM | Test Set |
|---|---|---|---|---|
| **Total number of studies** | 237 | 47 | 93 | 142 |
| **Total number of lesions** | 270 | 54 | 125 | 124 |
| **Size** (mean ± SD; median)<br>Max. diameter [mm]<br>Volume [mm$^3$] | 4.37 ± 2.73; 3.78<br>52.83 ± 155.03; 14.71 | 2.4 ± 1.01; 2.41<br>5.23 ± 4.14; 3.94 | 4.53 ± 2.19; 4.34<br>31.76 ± 48.66; 16.35 | 4.45 ± 3.49; 3.58<br>95.07 ± 388.65; 13.02 |

Table 1. Sizes of datasets and sizes and number of aneurysms in the different datasets.

The test set contained 124 UICA distributed over 101 positive studies and contained 41 negative studies without any UICA. Average lesion count across the test set was ~0.87/case, median maximum diameter was 3.58mm, and median volume was 13.02mm$^3$.

3.2) Performance of AP+DD+ADAM model

The AP+DD+ADAM model, which is the primary model of our study, demonstrated a sensitivity of 85% and an FP/case rate of 0.23 for UICA. For the correctly detected UICA it showed a high lesion-wise DICE score of 0.73 (measuring overlap accuracy across individual UICA). Furthermore, for any given correctly detected UICA, about 84% of the predicted segmentation surface remained within a margin of 0.5 mm to the ground truth (lesion-wise NSD = 0.84). Size difference between predictions and ground truth was on average 1.21mm (maximum diameter) and 59.52 mm$^3$ (volume), respectively.

3.3) Comparison of AP+DD+ADAM model with non-IH models

For UICA detection sensitivity, the AP+DD+ADAM model significantly outperformed the ADAM-winner (sensitivity: 61%; $p < 0.05$) and the ADAM-nnU-Net (sensitivity 51%; $p < 0.05$). There was no significant difference for the total FP/case despite the difference in sensitivity (ADAM-winner: 0.23; ADAM-nnU-Net: 0.21; p=0.88).

Regarding segmentation performance, there were no significant differences in DICE (0.73 vs 0.71, $p = 0.19$)) or NSD (0.84 vs 0.85, $p = 0.79$) between the AP+DD+ADAM and the ADAM-winner models. However, the AP+DD+ADAM model performed significantly better than the ADAM-nnU-Net in both DICE (0.73 vs 0.61, $p < 0.05$) and NSD (0.84 vs 0.74, $p < 0.05$).

The AP+DD+ADAM model and the ADAM-winner model reported no significantly different values regarding mean volume differences of the aneurysms (59.52 mm$^3$ vs 78.62 mm$^3$, p=0.68) and mean maximum diameter differences (1.21mm vs 2.08mm, p=0.06) between predictions and ground truth. The AP+DD+ADAM model significantly outperformed the ADAM-nnU-Net in both predicted mean volume differences (59.52 mm$^3$ vs 83.07 mm$^3$, p<0.05) and predicted mean maximum diameter differences (1.21mm vs 2.63mm, p<0.05) between predictions and ground truth.

3.4) Detection Performance of IH models

Detection performance metrics are summarized in Table 2, segmentation performance metrics are summarized in Table 3. Figure 2 shows image examples from selected patients.

|  | AP | AP+DD | AP+ADAM | AP+DD+ADAM | ADAM-nnU-Net | ADAM-winner |
|---|---|---|---|---|---|---|
| **Sensitivity [95%-CI]** | | | | | | |
| Across all aneurysms | 0.82 [0.76,0.89] | 0.85 [0.78,0.91] | 0.83 [0.76,0.90] | 0.85 [0.78,0.91] | 0.51 [0.42, 0.60] | 0.61 [0.53, 0.70] |
| aneurysms < 2mm | 0.40 [0.19,0.61] | 0.45 [0.23,0.67] | 0.45 [0.23,0.67] | 0.35 [0.14,0.56] | 0.15 [0.00, 0.31] | 0.15 [0.00, 0.31] |
| aneurysms < 4mm | 0.73 [0.62,0.83] | 0.76 [0.66,0.86] | 0.71 [0.61,0.82] | 0.74 [0.64,0.85] | 0.36 [0.24, 0.47] | 0.50 [0.38, 0.62] |
| aneurysms ≥ 4mm | 0.94 [0.88,1.00] | 0.96 [0.91,1.00] | 0.98 [0.95,1.00] | 0.98 [0.95,1.00] | 0.70 [0.58, 0.83] | 0.76 [0.65, 0.87] |
| **FP/case [95%-CI]** | | | | | | |
| Across all aneurysms | 0.20 [0.13, 0.26] | 0.31 [0.23, 0.38] | 0.24 [0.17, 0.31] | 0.23 [0.16, 0.3] | 0.21 [0.14, 0.28] | 0.23 [0.16, 0.3] |
| aneurysms < 2mm | 0.13 [0.08, 0.19] | 0.21 [0.14, 0.28] | 0.16 [0.10, 0.22] | 0.13 [0.07, 0.18] | 0.13 [0.07, 0.18] | 0.13 [0.07, 0.18] |
| aneurysms < 4mm | 0.18 [0.12, 0.25] | 0.30 [0.23, 0.38] | 0.21 [0.14, 0.28] | 0.22 [0.15, 0.28] | 0.20 [0.14, 0.27] | 0.21 [0.14, 0.28] |
| aneurysms ≥ 4mm | 0.01 [0, 0.03] | 0.01 [0, 0.02] | 0.03 [0, 0.05] | 0.01 [0, 0.03] | 0.01 [0, 0.02] | 0.02 [0, 0.04] |

Table 2: Summary of the different models' performance in both detection and segmentation on the test set.

| mean ± 95%-CI | AP | AP+DD | AP+ADAM | AP+DD+ADAM | ADAM-nnU-Net | ADAM-winner |
|---|---|---|---|---|---|---|
| **DICE** (correctly detected aneurysms) | 0.67 [0.62, 0.71] | 0.7 [0.66, 0.74] | 0.68 [0.63, 0.72] | 0.73 [0.69, 0.77] | 0.61 [0.54, 0.67] | 0.71 [0.66, 0.75] |
| **NSD (0.5mm)** (correctly detected aneurysms) | 0.81 [0.76, 0.85] | 0.84 [0.80, 0.88] | 0.81 [0.77, 0.85] | 0.84 [0.80, 0.88] | 0.74 [0.68, 0.80] | 0.85 [0.81, 0.89] |
| **Volume Differences** (correctly detected aneurysms) [mm$^3$] | 61.86 [13.51, 124.49] | 60.80 [12.49, 122.89] | 52.04 [9.67, 109.15] | 59.52 [10.89, 122.42] | 83.07 [24.49, 159.87] | 78.62 [19.45, 153.39] |
| **Maximum Diameter Differences** (correctly detected aneurysms) [mm] | 1.35 [1.04, 1.72] | 1.26 [0.95, 1.63] | 1.20 [0.95, 1.50] | 1.21 [0.89, 1.59] | 2.63 [2.05, 3.32] | 2.08 [1.49, 2.79] |
| **Spearman Correlation Coefficient for Volumes** (correctly detected aneurysms) | 0.88 [0.82, 0.91] | 0.90 [0.86, 0.93] | 0.94 [0.91, 0.96] | 0.92 [0.88, 0.94] | 0.71 [0.56, 0.82] | 0.91 [0.86, 0.94] |
| **Spearman Correlation Coefficient for Maximum Diameters** (correctly detected aneurysms) | 0.87 [0.81, 0.91] | 0.88 [0.82, 0.91] | 0.90 [0.86, 0.93] | 0.89 [0.85, 0.93] | 0.71 [0.56, 0.81] | 0.90 [0.84, 0.93] |

Table 3. Segmentation metrics (DICE, NSD) and mean size differences between true aneurysms and model predictions for the test set. Size differences are calculated as ground truth size – prediction size.

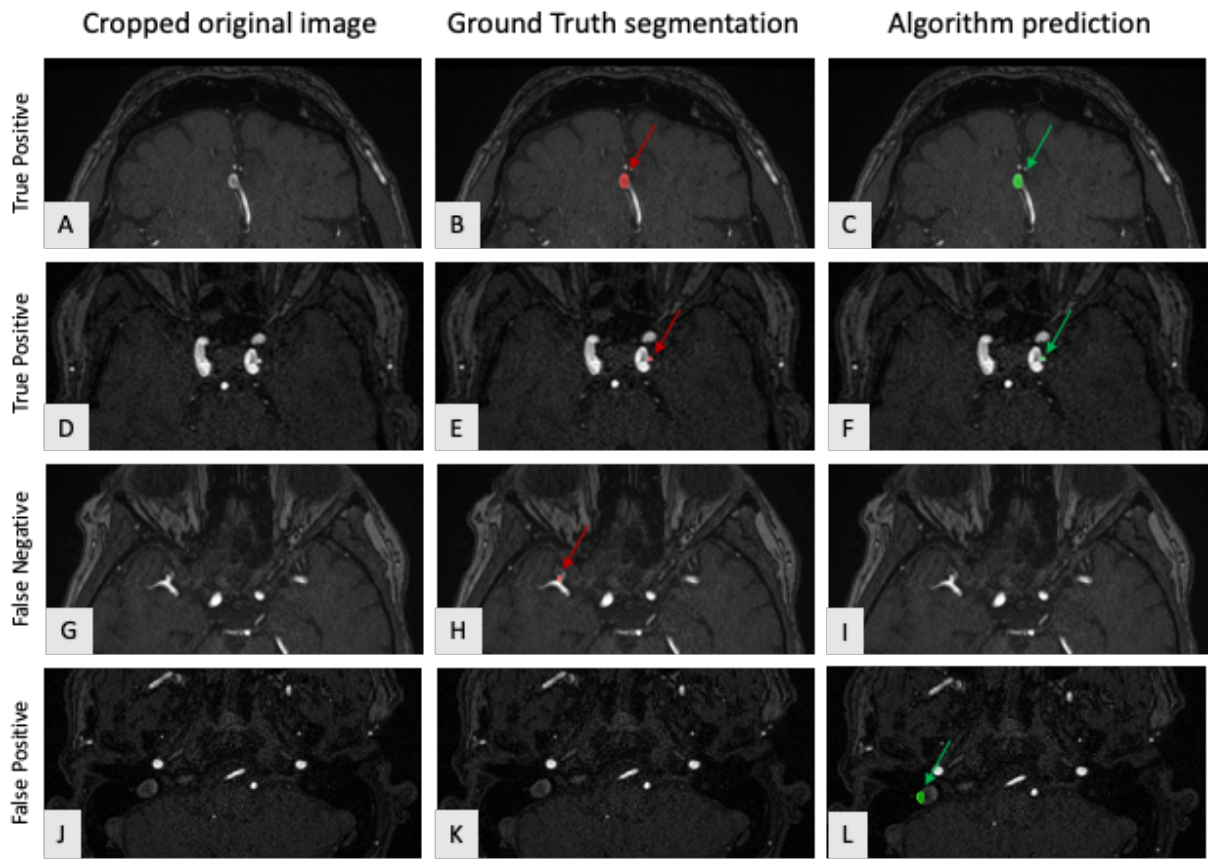

Figure 2. Examples of images, reader segmentations and the model predictions (AP+DD+ADAM model). A-C and D-F: Two examples of correctly identified aneurysms with corresponding segmentations. G-I: aneurysm missed by model. J-L: False Positive prediction in the internal jugular vein with no corresponding ground truth aneurysm.

IH models achieved an overall sensitivity ranging between 0.82 and 0.85 (no significant difference; p=0.94). Total FP/case rates of the IH models ranged from 0.20 to 0.31 (no significant difference; p=0.16). Ten aneurysms and eight aneurysm DD, which were not described in the clinical reports, were identified by the models and confirmed by the supervising neuroradiologists. These were then included as aneurysms in the datasets.

For IH models, sensitivity improved with increasing lesion diameter (Figure 3a), reaching 98% for UICA>4mm for the AP+DD+ADAM model (vs 0.74 for UICA <4mm, p<0.05). Correspondingly, the FP/case rate showed a marked decrease with increasing lesion diameter, with 0.01 for predictions >4mm (vs 0.22 for predictions <4mm, p<0.05) (Figure 3b).

3.5) Segmentation Performance of IH models

Mean lesion-wise DICE score showed no significant differences across IH models, varying between 0.67 and 0.73 (p=0.10). DICE score showed almost no dependency on lesion size (Figure 4). Mean lesion-wise NSD varied between 0.81 and 0.84 and showed no significant differences between IH models (p=0.33), meaning that for a given segmentation, between 81% and 84% remained in a margin of 0.5 mm to the ground truth.

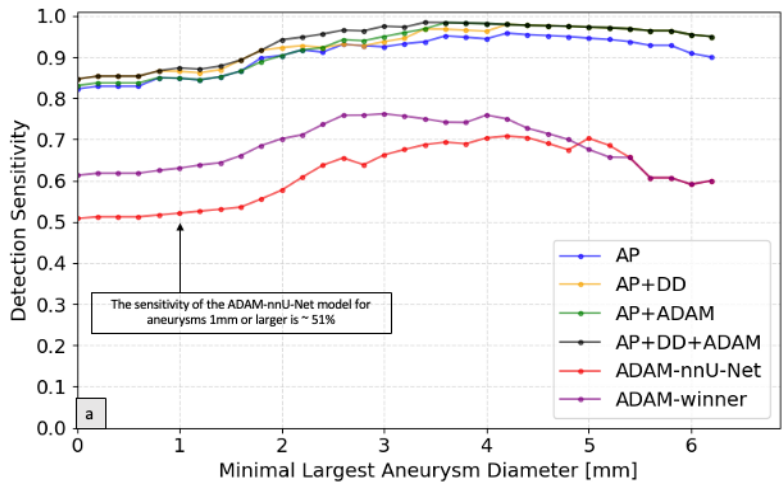

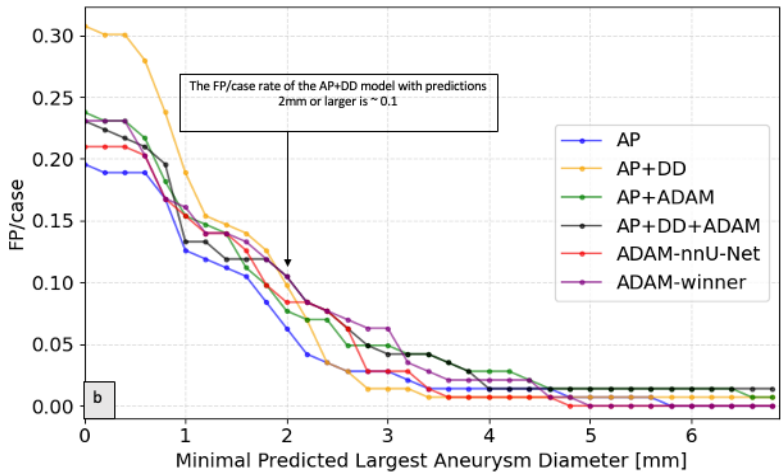

Figure 3: Diameter-dependent cumulative sensitivity (a) and FP/case rate (b) of all models. The figure shows detection sensitivity and FP/case for all detected aneurysms that are equal or larger than the diameter specified on the x-axis.

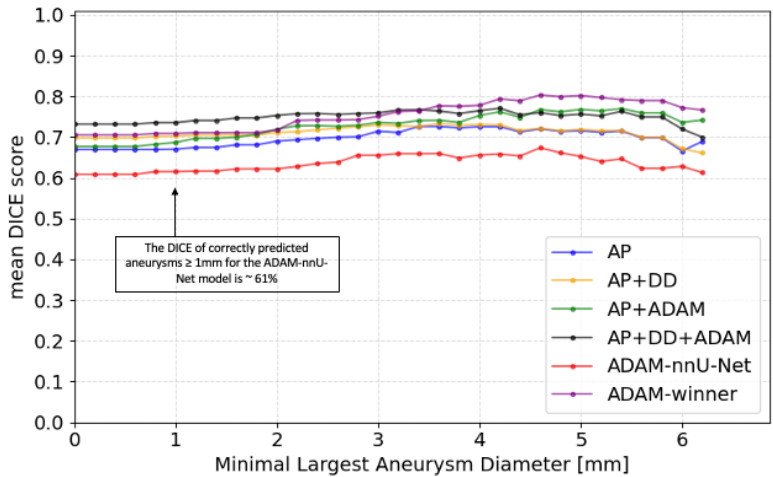

Figure 4: Diameter-dependent DICE score for all models. The figure shows mean DICE scores for all correctly detected UICA that are equal or larger than the diameter specified on the x-axis.

3.6) Ground Truth and Predicted Size Comparison

Comparison of the ground truth and predicted UICA sizes showed that for all IH models, the average volume of the predicted lesions was significantly smaller than the average volume of the ground truths. Spearman correlation coefficients between true and predicted sizes were between 0.88 and 0.94 for volumes and 0.87 and 0.90 for largest diameters and did not differ significantly (p = 0.07, p = 0.61). The average differences between ground truth and model predictions are listed in Table 3. Mean volume differences were between 52mm$^3$ and 61mm$^3$ for all IH models and mean maximum diameter differences were between 1.21mm and 1.35mm. There were no significant differences between the IH models for volume (p=0.75) and maximum diameter differences (p=0.73). Figure 5 visually depicts this relationship for the AP+DD+ADAM model.

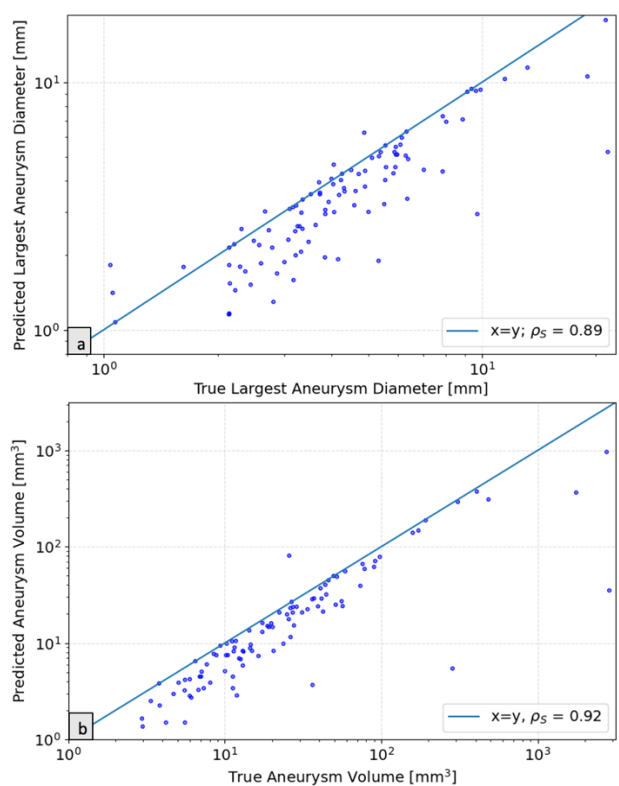

Figure 5: Correlation plots between true and predicted aneurysm largest diameters (a) and volumes (b) for the AP+DD+ADAM model.

## 4. Discussion

In our study, we introduce a deep-learning model based on the nnU-Net to address two major challenges in diagnosing UICAs on non-contrast 3D TOF-MRIs: combined detection and volumetric segmentation.

The first challenge, detecting UICAs in a routine clinical setting, is difficult due to increasing neuroradiologic workload and time constraints (Ivanovic et al., 2023). Many UICAs are found during unrelated imaging, where focusing the search pattern on small vessels requires considerable mental effort.

The model identified secondary aneurysms missed in initial readings, reducing satisfaction of search errors (Ivanovic et al., 2023) and improving detection of these not-so-rare cases (Rinne et al., 1994).

Our primary model trained on multi-center data detected aneurysms ≥4mm with a high sensitivity of 98% and a low false positive rate of 1 per 100 cases for findings ≥4mm. This is equal to the capabilities previously reported in humans (Sailer et al., 2014). UICA <4mm are less clinically significant but the model maintained a sensitivity of 74% and a low FP/case rate of 22 per 100 for such small findings. This size-dependent behavior is and probably mirrors human detection patterns.

Also, our metrics exceed previously reported models: Chen et al. (2023), Terasaki et al. (2021), Ueda et al. (2019) and Joo et al. (2021) all conducted multicentric studies for 3D aneurysm detection with good sensitivity (73%, 89.1%, 93%, 85.7%), but most had higher FP/case rates (0.88/case, 4.2/case, not provided for Ueda et al., 0.09/case). Importantly, none of these studies segmented aneurysms volumetrically.

We compared different models trained on different dataset combinations, some including clinically relevant and challenging uncertainty with potential differential diagnoses like arterial infundibula (Rutman et al., 2023). We did not find a significant difference for detection, segmentation and FP/case rate suggesting a remarkable capability and resilience in complex neurovascular cases.

Our model outperformed the ADAM challenge's winning model in sensitivity, likely due to our larger, more diverse training dataset with thorough expert segmentation. It also significantly surpassed a basic nnU-Net trained only on the ADAM dataset.

While the ADAM winner model was superior to the basic nnU-Net, probably due to loss function ensembling, it matched our primary model in segmentation quality. However, our model detected significantly more aneurysms, and those additional detections were segmented with equally high quality.

The second challenge in aneurysm diagnosis is accurately measuring aneurysm size and volume. Changes in size over time can indicate an increased risk of rupture, potentially necessitating further diagnostic or interventional procedures (Keedy et al., 2006). Therefore, precise aneurysm segmentation and volumetry with the extraction of maximal diameter and volume are relevant clinical features.

Considering the small size and complexity of aneurysms, our primary model demonstrated high accuracy in segmentation: It achieved an average DICE score of 0.73 for all correctly detected aneurysms. Additionally, a normalized surface distance of 0.84 indicates that 84% of the predicted segmentation surface is within one voxel of the ground truth, using a 0.5mm threshold on 3D TOF MRI scans with a slice thickness of 0.5mm. The model was able to reliably extract maximal diameter and volume of UICA with excellent correlation between ground truth and prediction. The first measure is integrated into the clinically relevant PHASES score (Greving et al., 2014) and the second offers a more comprehensive assessment of its true size and therefore its potential impact.

However, DICE-score was not size-dependent: For larger aneurysms, segmentation quality generally improved, but so did the incidence of aneurysms with inhomogeneous signal intensities (e.g., due to partial thrombosis or flow turbulences). The models struggled with such low-intensity regions, leading to poorer segmentations.

So far, only two models combined detection and volumetric segmentation of aneurysms in non-contrast TOF-MRIs: Sichtermann et al. (2019) had a sensitivity of 90% and a very high FP/case rate of 6.1, required extensive preprocessing and were evaluated with aneurysms almost double in size compared to our dataset. Claux et al. (2023) had a sensitivity of 78% and a FP/case rate of 0.5, both metrics being inferior to our model.

Our model could assist in assessing volumetric changes, where human performance is limited (Sahlein et al., 2023). However, this requires further validation is needed.

Our study has some limitations: While developed on multiple scanners with different field strengths and data from multiple imaging centers, it lacks further prospective evaluation in other centers. Further, we did not directly demonstrate non-inferiority compared to radiologists.

Our open-source nnU-Net-based AI model for UICA detection and segmentation on 3D TOF-MRI achieved 85% sensitivity and a low false positive rate of 0.23 per case. It also showed strong segmentation performance with a DICE score of 0.73 and an NSD of 0.84. This model could improve aneurysm detection in routine clinical settings and assist in monitoring aneurysm size over time. The model can be accessed open source under [10.5281/zenodo.13386859](https://doi.org/10.5281/zenodo.13386859) for research purposes.

# 5. References

1. Vlak, M. H., Algra, A., Brandenburg, R., & Rinkel, G. J. (2011). Prevalence of unruptured intracranial aneurysms, with emphasis on sex, age, comorbidity, country, and time period: A systematic review and meta-analysis. *The Lancet Neurology*, *10*(7), 626–636. https://doi.org/10.1016/s1474-4422(11)70109-0

2. Lantigua, H., Ortega-Gutierrez, S., Schmidt, J. M., Lee, K., Badjatia, N., Agarwal, S., Claassen, J., Connolly, E. S., & Mayer, S. A. (2015). Subarachnoid hemorrhage: Who dies, and why? *Critical Care*, *19*(1). https://doi.org/10.1186/s13054-015-1036-0

3. Keedy, A. (2006). An overview of intracranial aneurysms. *McGill Journal of Medicine*, *9*(2). https://doi.org/10.26443/mjm.v9i2.672

4. Lehnen, N.C., Schievelkamp, AH., Gronemann, C. et al. Impact of an AI software on the diagnostic performance and reading time for the detection of cerebral aneurysms on time of flight MR-angiography. Neuroradiology (2024). https://doi.org/10.1007/s00234-024-03351-w

5. Sohn, B., Park, K.-Y., Choi, J., Koo, J. H., Han, K., Joo, B., Won, S. Y., Cha, J., Choi, H. S., & Lee, S.-K. (2021). Deep learning–based software improves clinicians' detection sensitivity of aneurysms on Brain Tof-MRA. *American Journal of Neuroradiology*. https://doi.org/10.3174/ajnr.a7242

6. Isensee, F., Jaeger, P. F., Kohl, S. A., Petersen, J., & Maier-Hein, K. H. (2021). NNU-net: A self-configuring method for deep learning-based biomedical image segmentation. *Nature Methods*, *18*(2), 203–211. https://doi.org/10.1038/s41592-020-01008-z

7. Chen, Geng, Bao Yifang, Zhang Jiajun, Wang Dongdong, Zhou Zhiyong, Di Ruoyu, Dai Bin et al. "Automated unruptured cerebral aneurysms detection in TOF MR angiography images using dual-channel SE-3D UNet: a multi-center research." European Radiology 33, no. 5 (2023): 3532-3543.

8. Terasaki Y, Yokota H, Tashiro K, Maejima T, Takeuchi T, Kurosawa R, Yamauchi S, Takada A, Mukai H, Ohira K, Ota J, Horikoshi T, Mori Y, Uno T and Suyari H (2022) Multidimensional Deep Learning Reduces False-Positives in the Automated Detection of Cerebral Aneurysms on Time-Of-Flight Magnetic Resonance Angiography: A Multi-Center Study. Front. Neurol. 12:742126. doi: 10.3389/fneur.2021.742126

9. Ueda, D., Yamamoto, A., Nishimori, M., Shimono, T., Doishita, S., Shimazaki, A., Katayama, Y., Fukumoto, S., Choppin, A., Shimahara, Y., & Miki, Y. (2019). Deep learning for MR angiography: Automated detection of cerebral aneurysms. *Radiology*, *290*(1), 187–194. https://doi.org/10.1148/radiol.2018180901

10. Lehnen, N. C., Haase, R., Schmeel, F. C., Vatter, H., Dorn, F., Radbruch, A., & Paech, D. (2022). Automated Detection of Cerebral Aneurysms on TOF-MRA Using a Deep Learning Approach: An External Validation Study. *American Journal of Neuroradiology*, *43*(12), 1700–1705. https://doi.org/10.3174/ajnr.A7695

11. Claux, F., Baudouin, M., Bogey, C., & Rouchaud, A. (2023). Dense, deep learning-based intracranial aneurysm detection on TOF MRI using two-stage regularized U-Net. *Journal of Neuroradiology*, *50*(1), 9–15. https://doi.org/10.1016/j.neurad.2022.03.005

12. Sichtermann, T., Faron, A., Sijben, R., Teichert, N., Freiherr, J. & Wiesmann, M. (2019). Deep learning-based detection of intracranial aneurysms in 3D TOF-MRA. American Journal of Neuroradiology, 40(1), 25–32. https://doi.org/10.3174/ajnr.A5911

13. Timmins, K. M., van der Schaaf, I. C., Bennink, E., Ruigrok, Y. M., An, X., Baumgartner, M., Bourdon, P., de Feo, R., Noto, T. di, Dubost, F., Fava-Sanches, A., Feng, X., Giroud, C., Group, I., Hu, M., Jaeger, P. F., Kaiponen, J., Klimont, M., Li, Y.,Kuijf, H. J. (2021). Comparing methods of detecting and segmenting unruptured intracranial aneurysms on TOF-MRAS: The ADAM challenge. *NeuroImage*, *238*. https://doi.org/10.1016/j.neuroimage.2021.118216

14. Anastasopoulos C, Reisert M, Kellner E. "Nora Imaging": A Web-Based Platform for Medical Imaging. Neuropediatrics. 2017;48(S 01):P26.

15. JunMa11. (2020). ADAM2020 (Version v1.0) [Computer software]. Retrieved from https://github.com/JunMa11/ADAM2020.

16. Pauli Virtanen, Ralf Gommers, Travis E. Oliphant, Matt Haberland, Tyler Reddy, David Cournapeau, Evgeni Burovski, Pearu Peterson, Warren Weckesser, Jonathan Bright, Stéfan J. van der Walt, Matthew Brett, Joshua Wilson, K. Jarrod Millman, Nikolay Mayorov, Andrew R. J. Nelson, Eric Jones, Robert Kern, Eric Larson, CJ Carey, İlhan Polat, Yu Feng, Eric W. Moore, Jake VanderPlas, Denis Laxalde, Josef Perktold, Robert Cimrman, Ian Henriksen, E.A. Quintero, Charles R Harris, Anne M. Archibald, Antônio H. Ribeiro, Fabian Pedregosa, Paul van Mulbregt, and SciPy 1.0 Contributors. (2020) SciPy 1.0: Fundamental Algorithms for Scientific Computing in Python. Nature Methods, 17(3), 261-272.

17. Ivanovic V, Paydar A, Chang YM, Broadhead K, Smullen D, Klein A, Hacein-Bey L. Impact of Shift Volume on Neuroradiology Diagnostic Errors at a Large Tertiary Academic Center. Acad Radiol. 2023 Aug;30(8):1584-1588. doi: 10.1016/j.acra.2022.08.035. Epub 2022 Sep 27. PMID: 36180325.

18. Rinne J, Hernesniemi J, Puranen M, Saari T. Multiple intracranial aneurysms in a defined population: prospective angiographic and clinical study. Neurosurgery. 1994 Nov;35(5):803-8. doi: 10.1227/00006123-199411000-00001. PMID: 7838326.

19. Sailer AM, Wagemans BA, Nelemans PJ, de Graaf R, van Zwam WH. Diagnosing intracranial aneurysms with MR angiography: systematic review and meta-analysis. Stroke. 2014 Jan;45(1):119-26. doi: 10.1161/STROKEAHA.113.003133. Epub 2013 Dec 10. PMID: 24326447.

20. Joo B, Ahn SS, Yoon PH, Bae S, Sohn B, Lee YE, Bae JH, Park MS, Choi HS, Lee SK. A deep learning algorithm may automate intracranial aneurysm detection on MR angiography with high diagnostic performance. Eur Radiol. 2020 Nov;30(11):5785-5793. doi: 10.1007/s00330-020-06966-8. Epub 2020 May 30. PMID: 32474633.

21. Rutman AM, Wangaryattawanich P, Aksakal M, Mossa-Basha M. Incidental vascular findings on brain magnetic resonance angiography. Br J Radiol. 2023 Feb;96(1142):20220135. doi: 10.1259/bjr.20220135. Epub 2022 Apr 19. PMID: 35357891; PMCID: PMC9975521.

22. Greving, J. P., Wermer, M. J., Brown, R. D., Morita, A., Juvela, S., Yonekura, M., ... & Algra, A. (2014). Development of the PHASES score for prediction of risk of rupture of intracranial aneurysms: a pooled analysis of six prospective cohort studies. The Lancet Neurology, 13(1), 59-66.

23. Sahlein, D.H., Gibson, D., Scott, J.A., DeNardo, A., Amuluru, K., Payner, T., Rosenbaum-Halevi, D. and Kulwin, C., 2023. Artificial intelligence aneurysm measurement tool finds growth in all aneurysms that ruptured during conservative management. Journal of NeuroInterventional Surgery, 15(8), pp.766-770.

# 6. Supplementary Materials

| Scanner Model | Magnetic Field Strength [T] | $T_E$ [ms] | $T_R$ [ms] | Slice Thickness [mm] | # of studies |
|---|---|---|---|---|---|
| MAGNETOM Vida | 3.00 | 3.69 | 21.14 | 0.30 | 93 |
| Skyra | 3.00 | 3.42 | 21.00 | 0.50 | 90 |
| Skyra_fit | 3.00 | 3.42 | 21.00 | 0.50 | 75 |
| Avanto_fit | 1.50 | 7.00 | 23.00 | 0.50 | 40 |
| MAGNETOM Sola | 1.50 | 7.15 | 23.32 | 0.47 | 39 |
| Skyra | 3.00 | 3.43 | 21.00 | 0.50 | 15 |
| MAGNETOM Vida fit | 3.00 | 3.69 | 21.00 | 0.40 | 13 |
| Skyra | 3.00 | 3.42 | 19.00 | 0.60 | 5 |
| MAGNETOM Free.Max | 0.55 | 7.07 | 27.67 | 0.51 | 4 |
| MAGNETOM Vida | 3.00 | 3.69 | 21.18 | 0.30 | 3 |
| | | | 21.22 | 0.30 | 2 |
| Skyra | 3.00 | 3.19 | 20.00 | 0.70 | 2 |
| Skyra_fit | 3.00 | 3.42 | 19.00 | 0.60 | 1 |
| | | | | 0.60 | 1 |
| Skyra | 3.00 | 3.42 | 25.00 | 0.50 | 1 |
| MAGNETOM Sola | 1.50 | 7.00 | 22.40 | 0.60 | 1 |
| MAGNETOM Vida | 3.00 | 3.50 | 20.00 | 0.60 | 1 |
| Prisma | 3.00 | 3.42 | 21.00 | 0.50 | 1 |
| MAGNETOM Vida | 3.00 | 3.69 | 21.00 | 0.40 | 1 |
| | | 4.98 | 23.99 | 0.30 | 1 |
| | | 3.69 | 21.16 | 0.30 | 1 |
| | | | 21.14 | 0.40 | 1 |
| MAGNETOM Vida fit | 3.00 | 3.69 | 21.00 | 0.50 | 1 |

Supplement Table 1: Specifications of Scanners and Scanning Parameters used for TOF MRIs for ID.

| | Main Institute | Affiliate Center 1 | Affiliate Center 2 | Affiliate Center 3 |
|---|---|---|---|---|
| **% of Institutional Data** | 53% | 41% | 3% | 2% |

Supplement Table 2: Sources of TOF-MRIs for Institutional Data